\documentclass[pdflatex,sn-mathphys-num]{sn-jnl}

\usepackage{graphicx}%
\usepackage{multirow}%
\usepackage{amsmath,amssymb,amsfonts}%
\usepackage{amsthm}%
\usepackage{mathrsfs}%
\usepackage[title]{appendix}%
\usepackage{xcolor}%
\usepackage{textcomp}%
\usepackage{manyfoot}%
\usepackage{booktabs}%
\usepackage{algorithm}%
\usepackage{algorithmicx}%
\usepackage{algpseudocode}%
\usepackage{listings}%
\usepackage{tikz}
\usepackage{comment}

\theoremstyle{thmstyleone}%
\theoremstyle{thmstyletwo}%

\theoremstyle{thmstylethree}%

\begin{document}

\title[Hybrid quantum-classical neural network for sentiment analysis]{Hybrid quantum-classical neural network for sentiment analysis}


\author*[1]{\fnm{Giacomo} \sur{Cappiello}}\email{giaccap@imada.sdu.dk}

\author*[2]{\fnm{Filippo} \sur{Caruso}}\email{filippo.caruso@unifi.it}

\author*[3]{\fnm{Xing} \sur{Liang}}\email{x.liang@kingston.ac.uk}

\author*[3]{\fnm{Dimitrios} \sur{Makris}}\email{d.makris@kingston.ac.uk}

\affil[1]{\orgdiv{Center for Quantum Mathematics}, \orgname{University of Southern Denmark}, \orgaddress{\street{Campusvej 55}, \city{Odense}, \postcode{5230}, \country{Denmark}}}

\affil[2]{Dept. of Physics and Astronomy, Florence Univ., via Sansone 1, I-50019 Sesto Fiorentino, Italy}


\affil[3]{School of Computer Science and Mathematics, Kingston University London, KT1 2EE, United Kingdom}


\abstract{Quantum machine learning has recently emerged as a promising paradigm that leverages the expressive power of quantum circuits to address complex learning tasks. In this work, we investigate the applicability of hybrid quantum–classical neural networks to sentiment analysis, a central problem in natural language processing. We focus on a dataset of tweets related to COVID-19, where the textual content is vectorized using TF-IDF and fed into both classical feedforward networks and hybrid architectures incorporating parameterized quantum circuits. Our results show that hybrid models can achieve accuracy comparable to the classical baseline, while exhibiting distinct learning dynamics, especially in terms of validation loss and accuracy, that suggest a richer representational capacity. Moreover, when applying transfer learning to an SMS spam classification task, the hybrid models consistently outperform the classical counterpart, achieving an accuracy increase of 15 percentage points (from 66\% to 81\%) on the spam class, demonstrating enhanced generalization. These findings highlight the feasibility of employing QML for natural language processing and point toward the potential advantages of hybrid models as quantum hardware continues to advance.}

\keywords{Quantum machine learning, hybrid neural networks, natural language processing, sentiment analysis, COVID-19}



\maketitle

\section{Introduction}

Sentiment analysis is a foundational task in natural language processing (NLP), that deals with the automatic extraction and classification of subjective information such as emotions, opinions, and attitudes from text. Its importance is rapidly growing with the proliferation of user-generated content, which has made it possible to capture large-scale public sentiment in real time across diverse domains. Surveys such as \cite{pang2008opinion,liu2012sentiment} have highlighted applications in market analysis, customer feedback monitoring, and political forecasting. With the rise of social media, sentiment analysis has become increasingly relevant for tracking public reactions to events \cite{thelwall2011sentiment}, monitoring brand perception \cite{medhat2014sentiment}, and supporting crisis management during natural disasters or health emergencies \cite{saif2013semantic}. Twitter, in particular, has been widely used as a data source due to its high-volume, time-sensitive, and opinion-rich content \cite{pak2010twitter}. These properties make sentiment analysis an essential tool for understanding collective behavior, especially during rapidly evolving events such as the COVID-19 pandemic.

Traditional approaches of sentiment analysis techniques were initially informed by machine learning methods such as Naive Bayes, Support Vector Machines (SVMs), and maximum entropy models \cite{pang2002thumbs}. But these models often struggle to capture contextual cues or sarcasm and domain shifts. Deep learning architectures have since become the state of the art, with convolutional neural networks (CNNs) \cite{kim2014convolutional}, recurrent neural networks (RNNs) \cite{tang2015document} and transformer-based models \cite{vaswani2017attention} demonstrating superior capacity to learn complex semantic relationships. However these models typically require large training corpora and significant computational resources, raising questions about scalability, efficiency, and interpretability.

Quantum machine learning represents a promising alternative paradigm by exploiting the computational properties of quantum systems, including superposition, entanglement, and high-dimensional Hilbert spaces. Variational quantum circuits (VQCs) have emerged as a flexible tool for constructing parametrized quantum models that can be trained via gradient-based optimization \cite{cerezo2021variational}. Hybrid quantum–classical neural networks (HNNs) combine classical layers with quantum layers, enabling quantum models to be integrated into end-to-end trainable architectures \cite{schuld2019quantum}. Recent work highlights potential advantages of quantum models in expressivity and generalization \cite{abbas2021power}, supported by empirical demonstrations in classification tasks \cite{havlivcek2019supervised}, image recognition \cite{cong2019quantum}, anomaly detection \cite{amin2018quantum}, and more recently natural language processing \cite{wang2021quantumnlp}. Despite this growing interest, applications of QML to sentiment analysis remain limited, and the potential benefits of hybrid architectures in this domain are not yet fully understood.

Advancing research in sentiment analysis and hybrid quantum-classical neural networks is important for many reasons. In the context of sentiment analysis, improving model accuracy and robustness directly benefits applications in social media monitoring, public health, and political analysis. By investigating hybrid approaches in practical NLP tasks, we can better understand the conditions under which quantum components provide tangible benefits, guiding future algorithmic development and the design of quantum hardware.

In this work, we investigate the use of hybrid neural networks for sentiment classification of COVID-19-related tweets. We compare a classical feedforward neural network with hybrid architectures that incorporate parameterized quantum circuits with 6, 8, and 12 qubits. Classical features are encoded through angle embedding, followed by entangling operations and measurement of Pauli-$Z$ expectation values, which serve as inputs to subsequent classical layers. All quantum components are simulated classically. In addition to evaluating performance on the primary sentiment analysis task, we perform transfer learning experiments on a spam detection dataset to assess the generalization and adaptability of hybrid models across tasks.

The paper is structured as follows: Section \ref{sec:meth} discusses the tweet dataset on which our study is based and the methods used to extract features and classify them, via classical and hybrid models. The results are shown and discussed in Section \ref{sec:res}, which also includes considerations on the use of transfer learning. Finally, Section \ref{sec:con} presents conclusions and future research.

\section{Methods}\label{sec:meth}

\subsection{Dataset}

We base our study on a corpus of tweets related to the COVID-19 pandemic, manually annotated with sentiment labels. Each entry is categorized as \textit{positive}, \textit{neutral}, or \textit{negative}. The dataset is split into a training set of 41,159 tweets (18,046 positive, 15,398 negative, and 7,712 neutral) and a test set of 3,798 tweets (1,546 positive, 1,633 negative, and 619 neutral). Alongside the tweet text, the raw data includes metadata such as user location and timestamp, with identifying information anonymized to preserve privacy.  

For our experiments, we retain only the \texttt{Original Tweet} and \texttt{Label} fields. We apply standard preprocessing steps to reduce noise: lowercasing all text, removing URLs, punctuation, numbers, and non-alphabetic characters. The resulting clean text forms the basis for feature extraction.  

\begin{figure}[h]
\centering
\includegraphics[scale=.55]{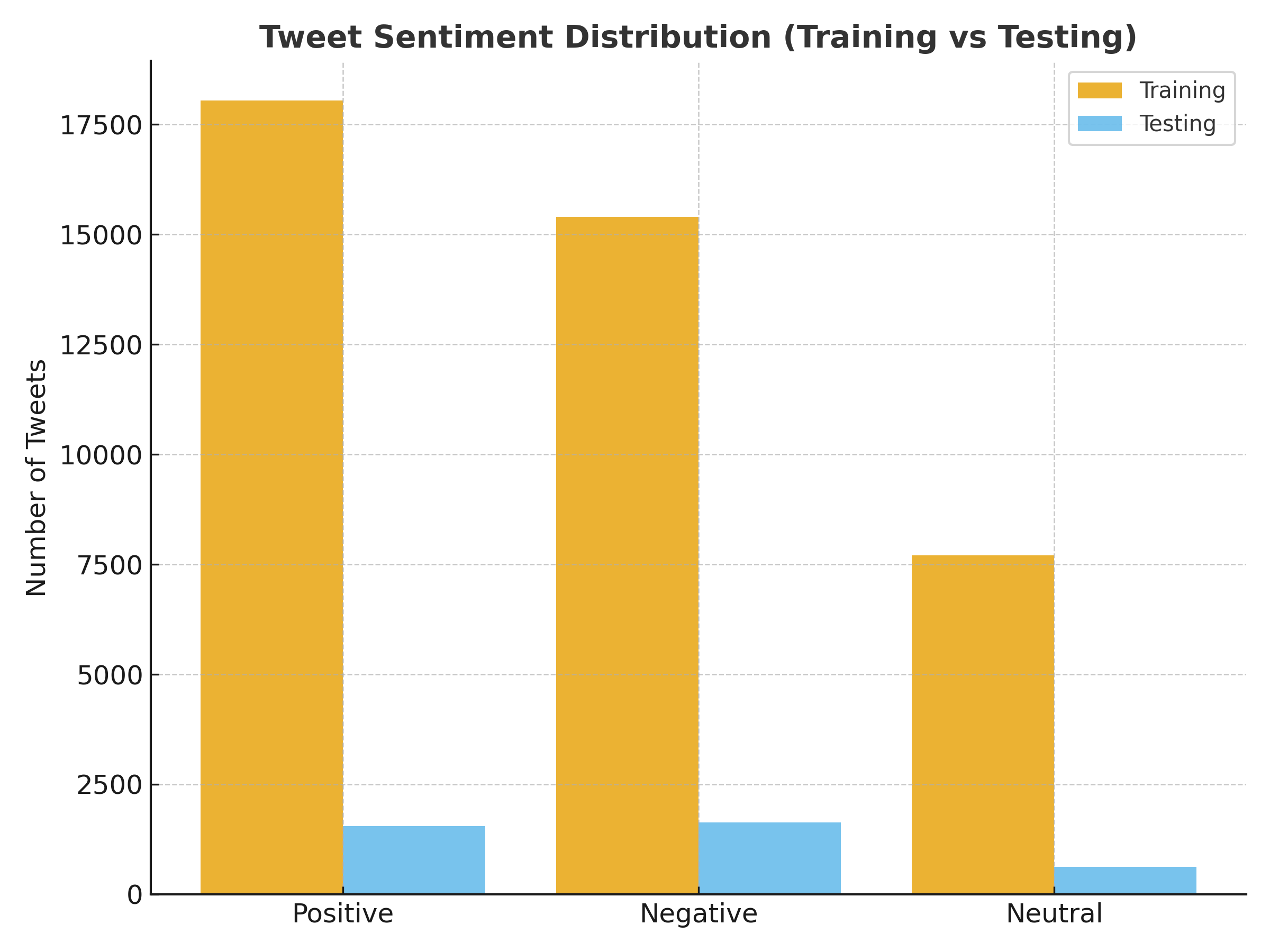}
\caption{Distribution of sentiment labels in the training and test sets.}
\label{fig:sent_distr}
\end{figure}

The sentiment distribution across training and test sets is illustrated in Fig.~\ref{fig:sent_distr}.  

\subsection{Feature extraction}

To represent the tweets in a form suitable for machine learning models, we rely on the Term Frequency-Inverse Document Frequency (TF-IDF) vectorization technique \cite{ramos2003using}. TF-IDF maps each document to a fixed-length vector by weighting term frequencies against their global distribution in the corpus. In this way, common words that carry little information are downweighted, while more distinctive terms are emphasized. This approach is widely adopted in text classification tasks \cite{joachims1998text} and provides a strong and efficient baseline for sentiment analysis.  

For comparison, we also experimented with Word2Vec embeddings \cite{mikolov2013efficient}, which encode semantic relationships in a continuous vector space. However, given the brevity and noisiness of tweets, Word2Vec did not improve performance, likely due to the lack of sufficient contextual information in short messages.  

In our experiments, the \texttt{TfidfVectorizer} was configured with the following parameters: \texttt{min\_df=1}, to include all words appearing at least once; \texttt{max\_df=0.95}, to exclude extremely frequent words; and \texttt{max\_features=5000}, to retain the 5,000 most informative terms. The vectorizer was first fit on the training set using \texttt{fit\_transform}, producing sparse vectors of dimension 5,000 for each tweet. The same transformation was then applied to the test set using \texttt{transform}, ensuring consistency across datasets.

\subsection{Neural Network}

Artificial neural networks (ANNs) are a family of computational models designed to approximate complex functions by composing layers of interconnected units, or neurons \cite{bishop2006pattern}. Each neuron computes a weighted sum of its inputs followed by a nonlinear activation, enabling the network to capture non-linear relationships in data. Feedforward neural networks, where information flows unidirectionally from the input to the output layer, represent the simplest and most widely used architecture \cite{goodfellow2016deep}. Modern networks are typically trained using stochastic gradient descent and its variants \cite{livni2014computational}, with regularization techniques such as dropout to improve generalization \cite{srivastava2014dropout}.

In this work, we establish a baseline for sentiment classification using a feedforward neural network applied to TF-IDF features. Each tweet is represented as a sparse 5,000-dimensional vector, which serves as the input to the model. The network architecture consists of:
\begin{itemize}
    \item an input layer of size 5,000;
    \item one hidden fully connected layer with 256 neurons, followed by a ReLU activation \cite{nair2010rectified};
    \item a dropout layer with a rate of 0.3 to mitigate overfitting;
    \item an output layer with 3 neurons, corresponding to the sentiment classes (\textit{positive}, \textit{neutral}, \textit{negative}), followed by a softmax activation to produce class probabilities \cite{asadi2020approximation}.
\end{itemize}

This simple design reduces the dimensionality of the TF-IDF representation while enabling the model to learn compact latent features. The resulting feedforward neural network provides a strong classical baseline for comparison with hybrid quantum architectures.

\subsection{Hybrid Neural Networks}

\begin{figure}[h!]
\centering
\includegraphics[]{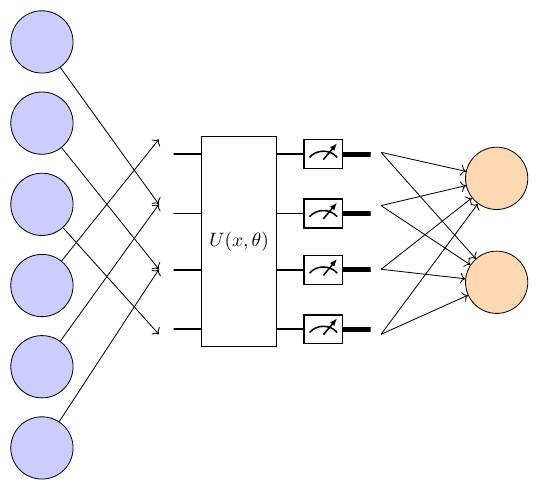}
\caption{Schematic representation of a hybrid quantum-classical neural network. Classical data is encoded into quantum states, processed through a PQC, and measured outputs are passed to classical layers.}
\label{fig:hybrid_nn}
\end{figure}

Hybrid Neural Networks (HNNs) integrate quantum circuits as modular components within classical neural network architectures, enabling end-to-end training that exploits both classical and quantum resources \cite{mitarai2018quantum, schuld2019evaluating, abbas2021power}. Classical data is encoded into quantum states via parameterized quantum circuits (PQCs), which act as non-linear feature extractors. The outputs of quantum measurements, typically expectation values of selected observables, are then fed into classical layers, allowing the network to jointly optimize classical and quantum parameters in a fully differentiable framework (see Fig. \ref{fig:hybrid_nn}).

In quantum computing, the basic unit of information is the \emph{qubit}, whose state resides in a two-dimensional Hilbert space $\mathcal{H}_2$ \cite{Nielsen_Chuang_2010}. An arbitrary qubit state can be written as
\begin{equation}
|\psi\rangle = \alpha |0\rangle + \beta |1\rangle, \quad |\alpha|^2 + |\beta|^2 = 1,
\end{equation}
or equivalently as
\begin{equation}
|\psi\rangle = \cos\frac{\theta}{2}|0\rangle + e^{i\varphi}\sin\frac{\theta}{2}|1\rangle.
\end{equation}

Single-qubit gates correspond to matrices in $\mathcal{U}(2)$, the group of $2\times2$ unitaries, including the Pauli gates:
\begin{equation}
X = \begin{bmatrix} 0 & 1 \\ 1 & 0 \end{bmatrix}, \;
Y = \begin{bmatrix} 0 & -i \\ i & 0 \end{bmatrix}, \;
Z = \begin{bmatrix} 1 & 0 \\ 0 & -1 \end{bmatrix},
\end{equation}
the Hadamard gate:
\begin{equation}
H = \frac{1}{\sqrt{2}}\begin{bmatrix} 1 & 1 \\ 1 & -1 \end{bmatrix},
\end{equation}
and rotation gates:
\begin{equation}
R_x(\theta) = \begin{bmatrix} \cos\frac{\theta}{2} & -i\sin\frac{\theta}{2} \\ -i\sin\frac{\theta}{2} & \cos\frac{\theta}{2} \end{bmatrix}, \quad
R_y(\theta) = \begin{bmatrix} \cos\frac{\theta}{2} & -\sin\frac{\theta}{2} \\ \sin\frac{\theta}{2} & \cos\frac{\theta}{2} \end{bmatrix}, \quad
R_z(\theta) = \begin{bmatrix} e^{-i\theta/2} & 0 \\ 0 & e^{i\theta/2} \end{bmatrix}.
\end{equation}

Multi-qubit controlled gates introduce entanglement, which is crucial for capturing correlations between features \cite{schuld2019quantum}. The controlled-NOT ($CX$) gate is defined as
\begin{equation}
CX = \begin{bmatrix}
1 & 0 & 0 & 0\\
0 & 1 & 0 & 0\\
0 & 0 & 0 & 1\\
0 & 0 & 1 & 0
\end{bmatrix},
\end{equation}
while the controlled-Z ($CZ$) gate is
\begin{equation}
CZ = \begin{bmatrix}
1 & 0 & 0 & 0\\
0 & 1 & 0 & 0\\
0 & 0 & 1 & 0\\
0 & 0 & 0 & -1
\end{bmatrix}.
\end{equation}

In this work, the HNN is applied to sentiment analysis using TF-IDF representations of tweets. The classical front-end consists of two fully connected layers: the first reduces the $5000$-dimensional TF-IDF vectors to a $256$-dimensional latent space with ReLU activation and dropout for regularization, while the second layer further compresses the features to match the number of qubits in the quantum circuit, with three configurations considered: $6$, $8$, or $12$ qubits. Classical features are embedded into the quantum layer via angle encoding, where each input controls the rotation of a single qubit. The qubits are then entangled through controlled operations, and measurements in the computational basis produce expectation values of Pauli-$Z$ observables, which serve as a non-linear transformation of the input features. Finally, these quantum outputs are passed through a classical linear layer to map them to the three sentiment classes, with softmax activation producing class probabilities.

Training is performed in a hybrid quantum-classical loop. Gradients of classical parameters are computed via standard backpropagation \cite{rumelhart1986learning}, while gradients of quantum parameters are obtained using the parameter-shift rule \cite{schuld2019evaluating}, enabling end-to-end optimization. This hybrid design combines the expressive capabilities of quantum circuits with the scalability of classical neural networks, allowing the model to capture complex patterns in textual data while remaining trainable on current noisy intermediate scale quantum (NISQ) hardware.

\subsection{Implementation}

The quantum layer is implemented as PQC, within a Pennylane \cite{PennyLane2023} \texttt{QNode}, acting on $Q$ qubits, with $Q$ taking the values $6$, $8$, or $12$. Classical input features are encoded into the quantum state using \emph{angle embedding}, where each feature determines the rotation angle of an $R_x$ gate applied to a corresponding qubit, effectively mapping the input vector into the quantum amplitudes. 

The circuit then applies two layers of the \texttt{BasicEntanglerLayers} template. Each layer combines parameterized single-qubit rotations with entangling operations across the qubits, introducing trainable parameters that allow the quantum circuit to adapt during training while maintaining qubit entanglement.  

Measurements of the Pauli-$Z$ operator are performed on each qubit at the end of the circuit, producing $Q$ outputs. These expectation values serve as the output of the quantum layer and are fed into subsequent classical layers in the hybrid neural network. All quantum components are executed in a noiseless classical simulation environment.

To ensure a fair comparison between classical and hybrid architectures, both follow the same training pipeline. Each experiment is repeated over $5$ independent runs with different random seeds to account for stochastic variations in optimization and data splitting. Within each run, $15\%$ of the training set is set aside as a validation subset for model selection.  

Training is performed for $30$ epochs, using the cross-entropy loss as the objective function \cite{mao2023cross}. To address class imbalance, loss weights are computed from the inverse frequency of each sentiment class. Optimization employs the \textit{Adam} \cite{kingma2014adam} with a learning rate of $10^{-4}$ and a weight decay of $10^{-5}$ for improved generalization. A StepLR scheduler reduces the learning rate by half every $20$ epochs to facilitate convergence, and mini-batches of size $64$ are used throughout.  

The model state achieving the lowest validation loss is retained for evaluation on the test set, and performance is reported as the average test accuracy over the $5$ runs.

\section{Results}\label{sec:res}

\begin{figure}[H]
    \centering
    \includegraphics[scale=.6]{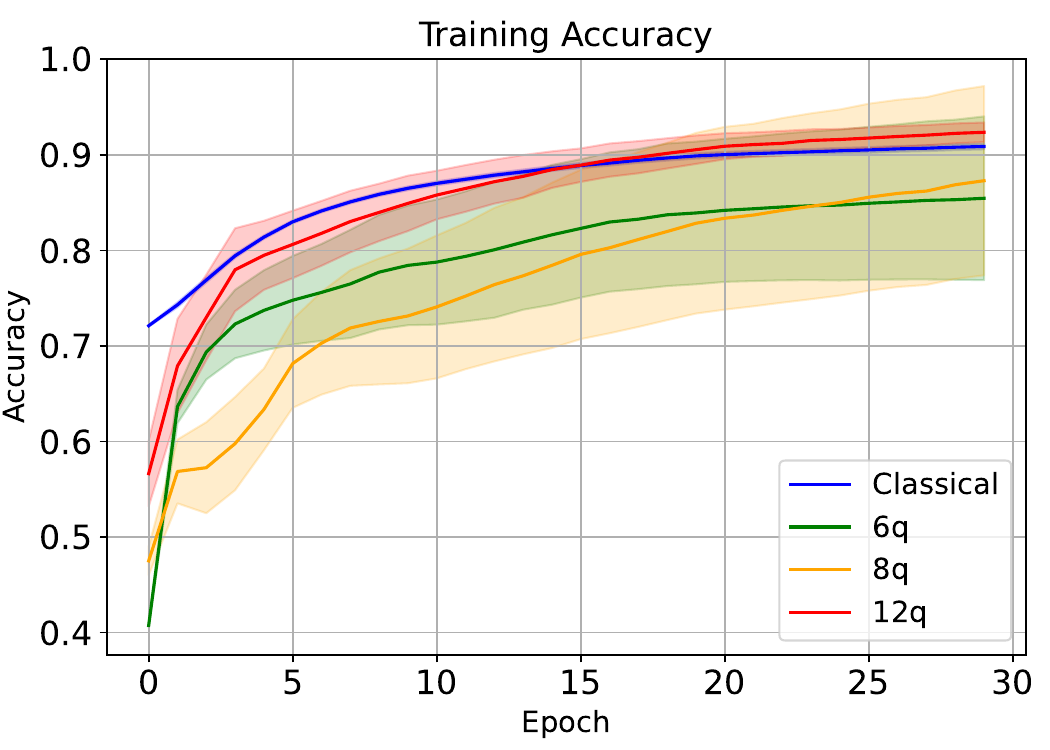}
    \includegraphics[scale=.6]{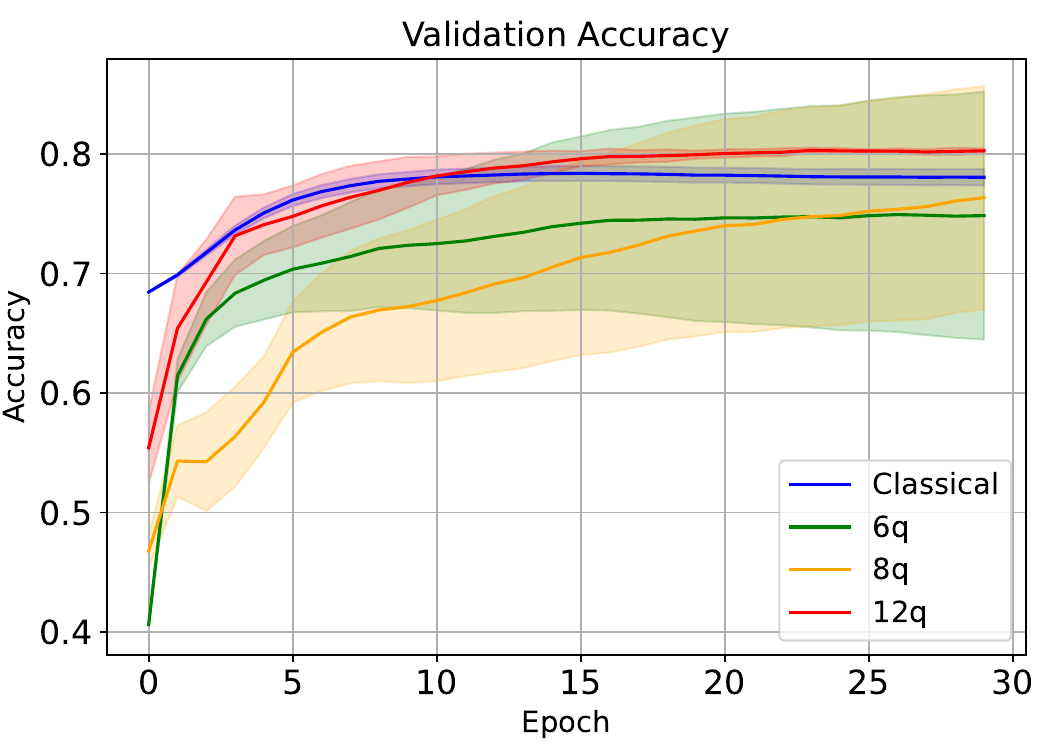}
    \caption{Mean and standard deviation of training and validation accuracy over 5 independent runs across 30 epochs, for the classical neural network and the hybrid neural networks with 6-, 8-, and 12-qubit quantum layers.}
    \label{fig:wo1}
\end{figure}

\begin{figure}[H]
    \centering
    \includegraphics[scale=.6]{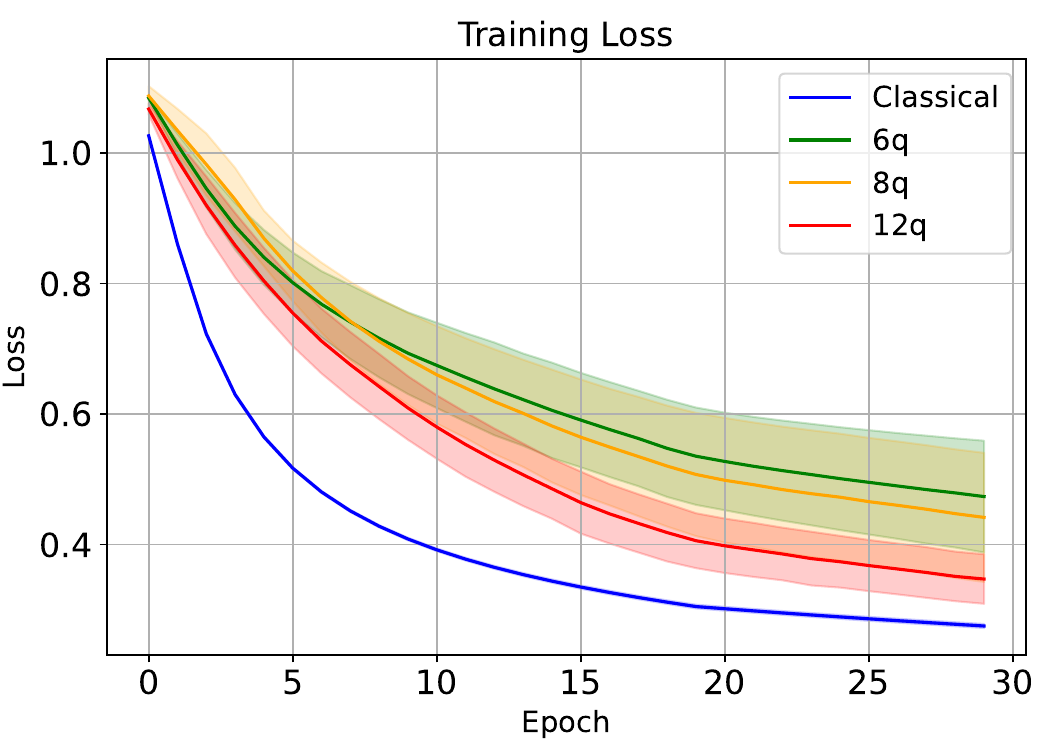}
    \includegraphics[scale=.6]{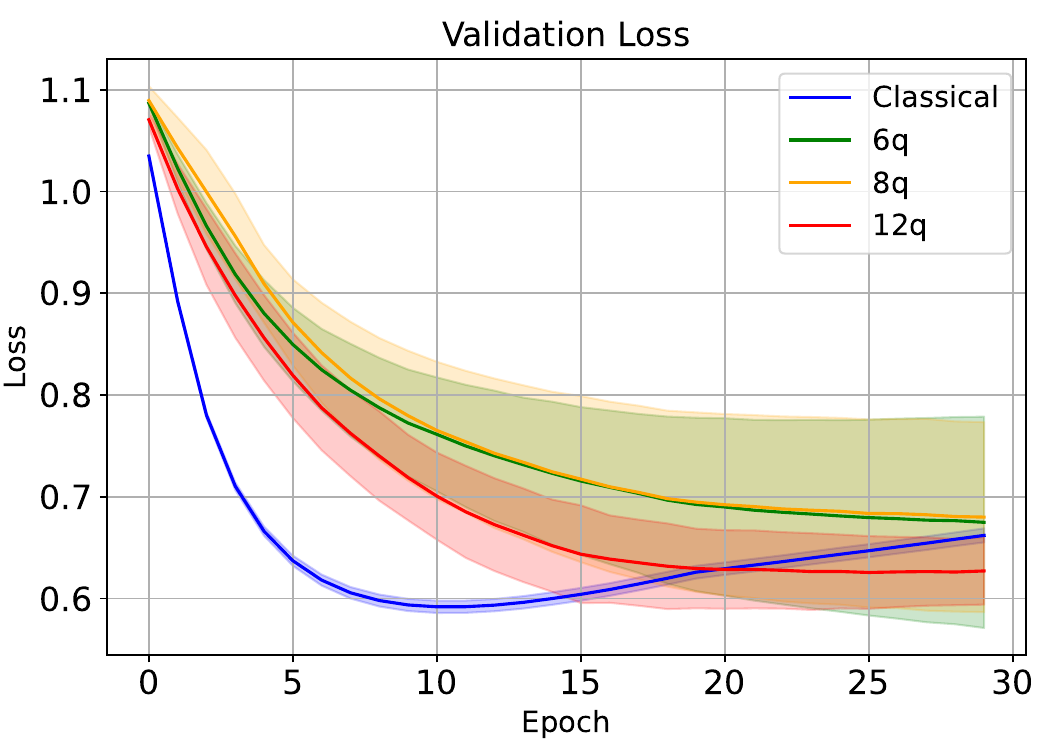}
    \caption{Mean and standard deviation of training and validation loss over 5 independent runs across 30 epochs, for the classical neural network and the hybrid neural networks with 6-, 8-, and 12-qubit quantum layers.}
    \label{fig:wo2}
\end{figure}

The mean and standard deviation of both training and validation losses, as well as training and validation accuracies, are illustrated in Fig. \ref{fig:wo1} and \ref{fig:wo2}. The figures present the results over $5$ independent runs, each consisting of $30$ training epochs, for the classical neural network and the hybrid neural networks with $6$-, $8$-, and $12$-qubit quantum layers.

From the plots, it is evident that the classical neural network exhibits the most consistent behavior across runs, with relatively low variance in both accuracy and loss. On the other hand, the $12$-qubit HNN achieves the highest training and validation accuracy and the lowest validation loss at the end of the $30$ epochs. The other hybrid models show higher variability, with substantially larger standard deviations across all metrics and overall lower performance. Notably, the classical model reaches its minimum validation loss early, around epoch $10$, but the loss subsequently begins to increase, suggesting potential early overfitting. In contrast, the hybrid networks continue to reduce the validation loss throughout training, indicating that the more complex architecture allows them to capture data patterns more effectively over time.

Table \ref{tab:wo} reports the mean and standard deviation of the test accuracy for the classical and hybrid models across $5$ runs. 

\begin{table}[h]
\caption{Mean accuracy and standard deviation (expressed in percentage) over 5 runs of the classical neural network and hybrid neural networks with 6-, 8-, and 12-qubit quantum layers.}
\begin{tabular*}{\textwidth}{@{\extracolsep\fill}cccc}
\toprule
\textbf{Classical} & & \textbf{Hybrid} & \\
\midrule
& 6q & 8q & 12q  \\ \midrule
$77.88\pm 0.25$ & $76.90\pm4.09$ & $75.92\pm 5.18$ & $77.64 \pm 1.26$\\\bottomrule
\end{tabular*}
\label{tab:wo}
\end{table}

The classical neural network achieves the highest average test accuracy, closely followed by the $12$-qubit hybrid model. Both models demonstrate strong performance, with low variability across runs, suggesting robust generalization and stable results. The $6$- and $8$-qubit hybrid networks show higher variance, indicating sensitivity to initialization and stochastic optimization.

To further analyze the classification behavior, confusion matrices are computed by comparing the predicted class labels with the true labels on the test set. Each element $(i,j)$ of the matrix represents the percentage of instances whose true label is $i$ and that are predicted as $j$. In this study, class labels are numeric: $0$ for negative sentiment, $1$ for neutral sentiment, and $2$ for positive sentiment. Confusion matrices thus allow assessment of not only overall accuracy but also the specific types of errors the models tend to make, such as confusing neutral tweets with positive or negative sentiment.

\begin{figure}[h!]
  \centering

  \includegraphics[scale=.6]{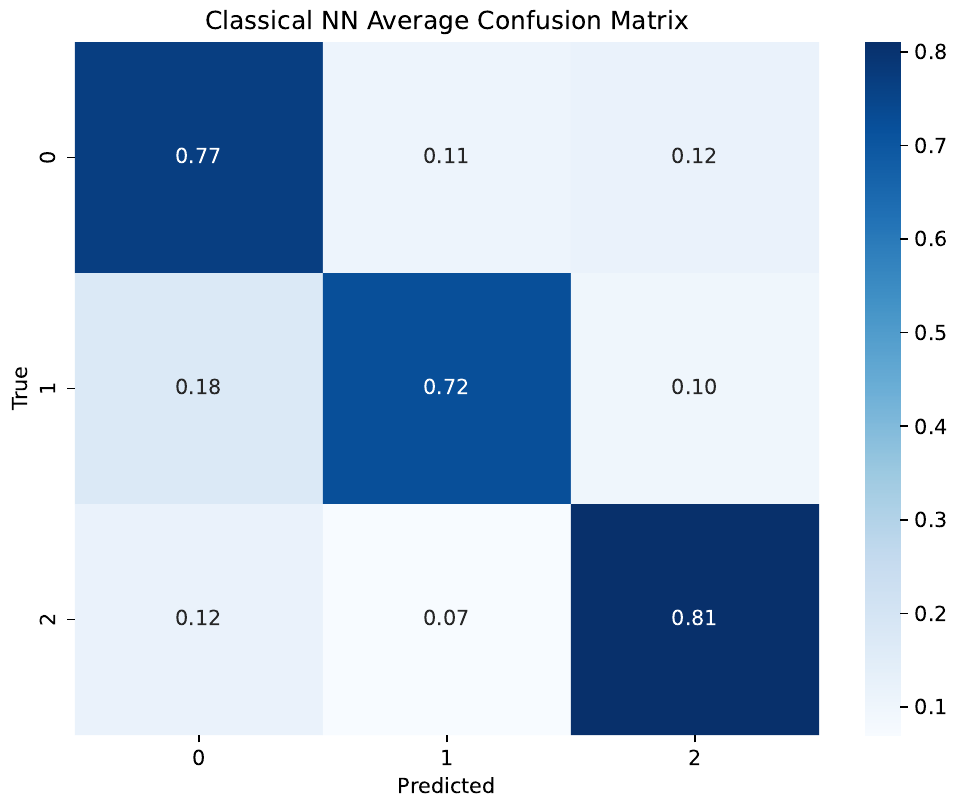}
  \includegraphics[scale=.6]{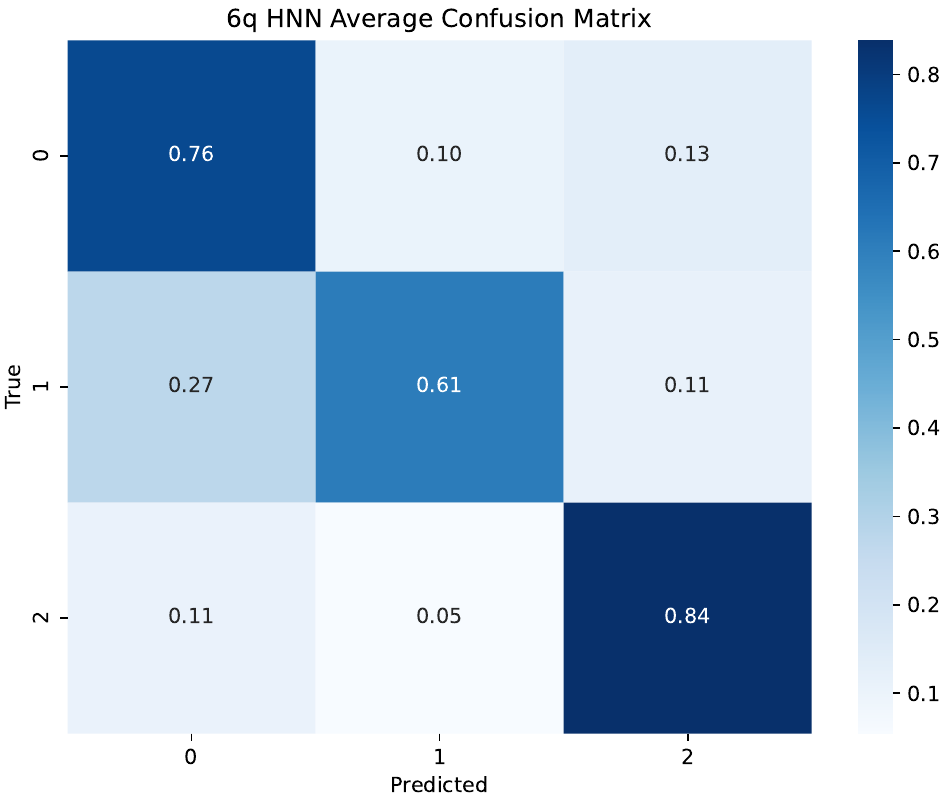}

  \caption{Average confusion matrices over 5 runs of the classical and hybrid neural network with 6-qubit quantum layer.}
  \label{fig:confu1}
\end{figure}

\begin{figure}[h!]
  \centering

  \includegraphics[scale=.6]{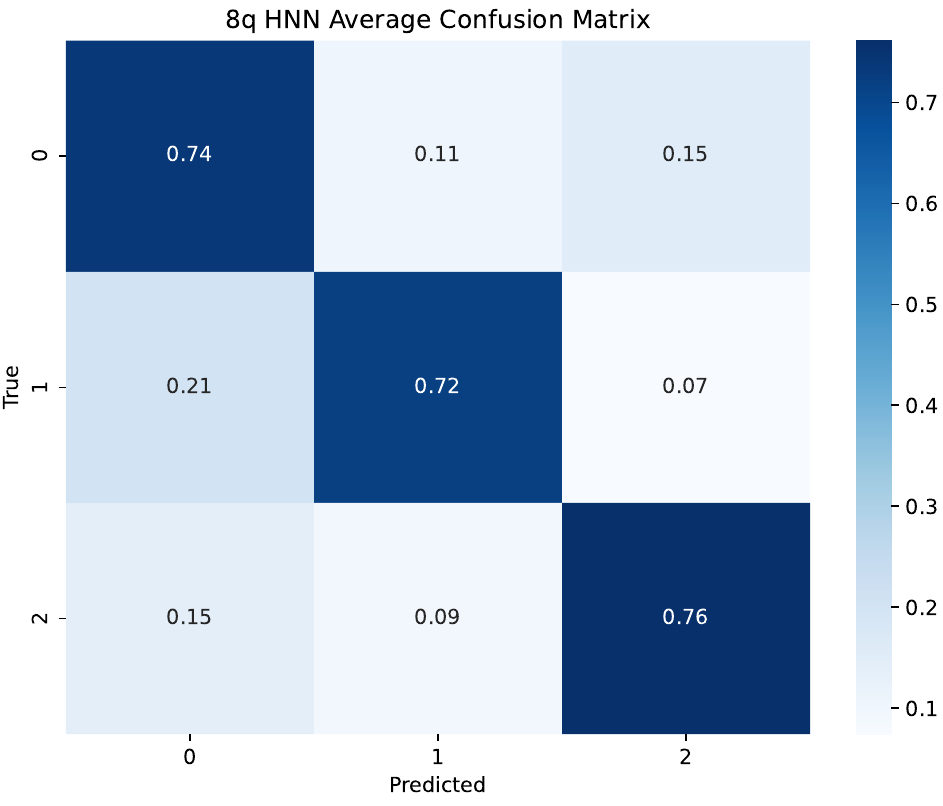}
  \includegraphics[scale=.6]{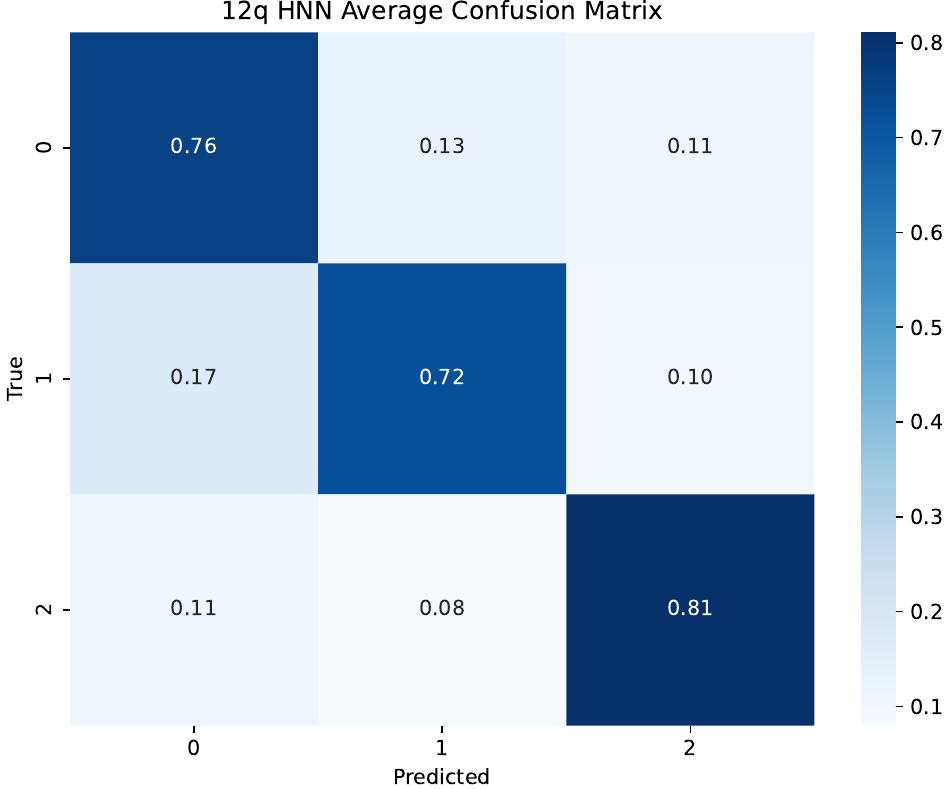}

  \caption{Average confusion matrices over 5 runs of the hybrid neural networks with 8- and 12-qubit quantum layers.}
  \label{fig:confu2}
\end{figure}

The average confusion matrices over the $5$ runs are presented in Fig. \ref{fig:confu1} and \ref{fig:confu2}. Both the classical model and the $12$-qubit HNN display balanced and reliable classification, whereas the $6$- and $8$-qubit hybrid models show different trade-offs. In particular, the $8$-qubit network exhibits reduced performance on the positive sentiment class, while the $6$-qubit model improves performance on the positive class at the expense of neutral sentiment predictions.

\subsection{Transfer learning}

Transfer learning enables models to exploit knowledge gained from one task to enhance performance on a related but distinct task \cite{pan2009survey}. In sentiment analysis, it is particularly useful when labeled data is limited or when training a model from scratch would be computationally expensive. By leveraging features learned on a larger or more complex dataset, the model can quickly adapt to new domains, such as different writing styles, topics, or labeling schemes.

In this case we employ transfer learning in a different way: we train a portion of the layers of our networks on the COVID tweets set and test the models on a new dataset, to evaluate their ability to adapt to a different task. Therefore we do not compare the performance of the networks with and without transfer learning, but we just confront the accuracy reached by the classical method with the one achieved by the hybrid ones.

For this study, we employ the \textit{SMS Spam Collection Dataset}, which contains $5574$ messages labeled as either \textit{ham} (legitimate) or \textit{spam}. This dataset differs from the original COVID-related sentiment dataset in size and task formulation, providing a suitable scenario to evaluate the transferability of features learned by the neural networks.

The neural network architectures are adapted for binary classification. In both classical and hybrid models, the original output layer of $3$ neurons (for negative, neutral, and positive sentiment) is replaced by an output layer of $2$ neurons, corresponding to ham and spam. Hybrid models are tested with $6$-, $8$-, and $12$-qubit quantum layers.

The transfer learning procedure can be divided in two stages:
\begin{itemize}
    \item \textit{pretraining:} the first two layers of both classical and hybrid models are pretrained on the COVID sentiment training set. These pretrained layers are then frozen.
    \item \textit{fine-tuning:} for the classical model, the output layer is trained on $80\%$ of the SMS spam dataset. For the hybrid models, both the quantum layer and the output layer are trained on the same subset. The remaining $20\%$ of the SMS dataset is reserved for testing.
\end{itemize}

Table \ref{tab:trans} reports the mean and standard deviation of test accuracies over $5$ runs after transfer learning.

\begin{table}[h!]
\caption{Mean accuracy and standard deviation (expressed in percentage) over 5 runs of classical and hybrid neural networks after transfer learning on the SMS spam dataset, with hybrid models using 6-, 8-, and 12-qubit quantum layers.}
\begin{tabular*}{\textwidth}{@{\extracolsep\fill}cccc}
\toprule
\textbf{Classical} & & \textbf{Hybrid} & \\
\midrule
& 6q & 8q & 12q  \\ \midrule
$63.12\pm 0.91$ & $65.08\pm1.28$ & $66.01\pm 3.06$ & $66.94\pm 1.85$
\\\bottomrule
\end{tabular*}
\label{tab:trans}
\end{table}

The results show that hybrid models consistently outperform the classical baseline in both overall accuracy and class-specific performance, indicating that the quantum layer enhances the network’s ability to capture informative patterns. 

\begin{figure}[H]
  \centering

  \includegraphics[scale=.55]{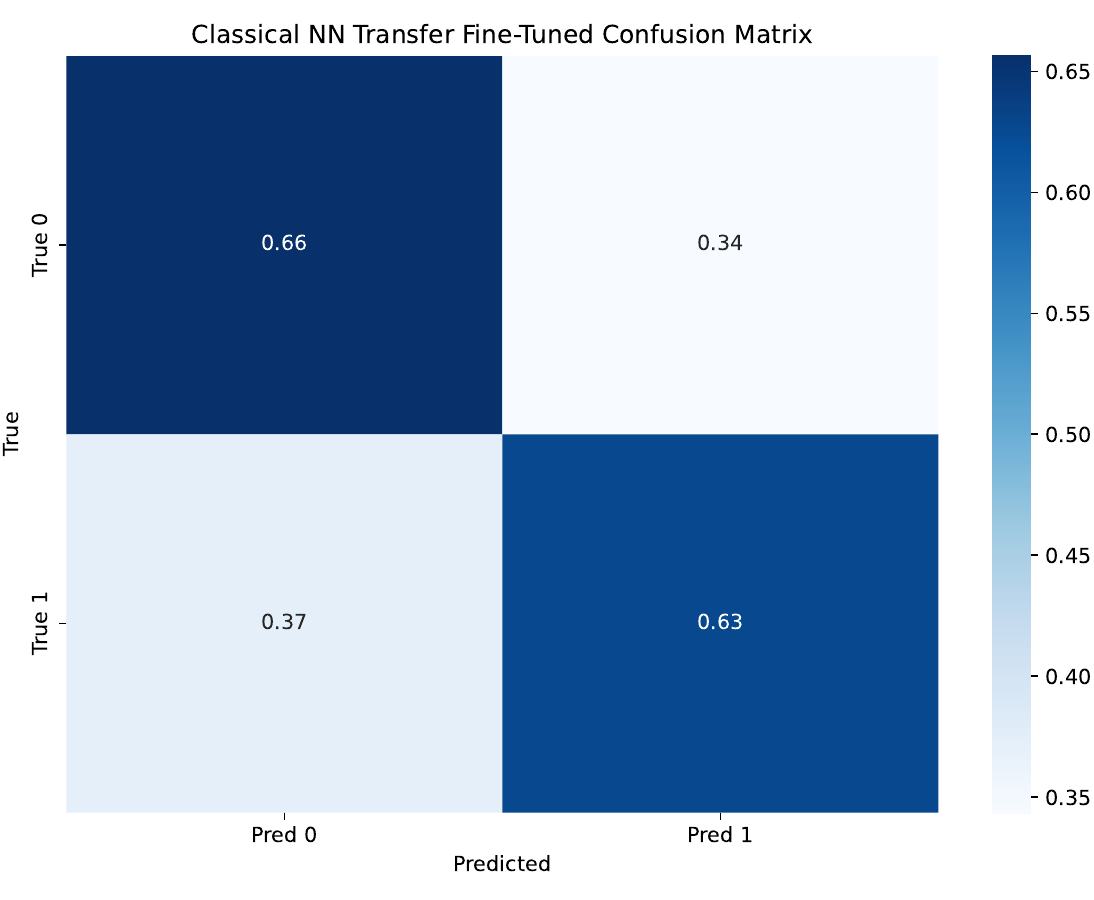}

  \includegraphics[scale=.55]{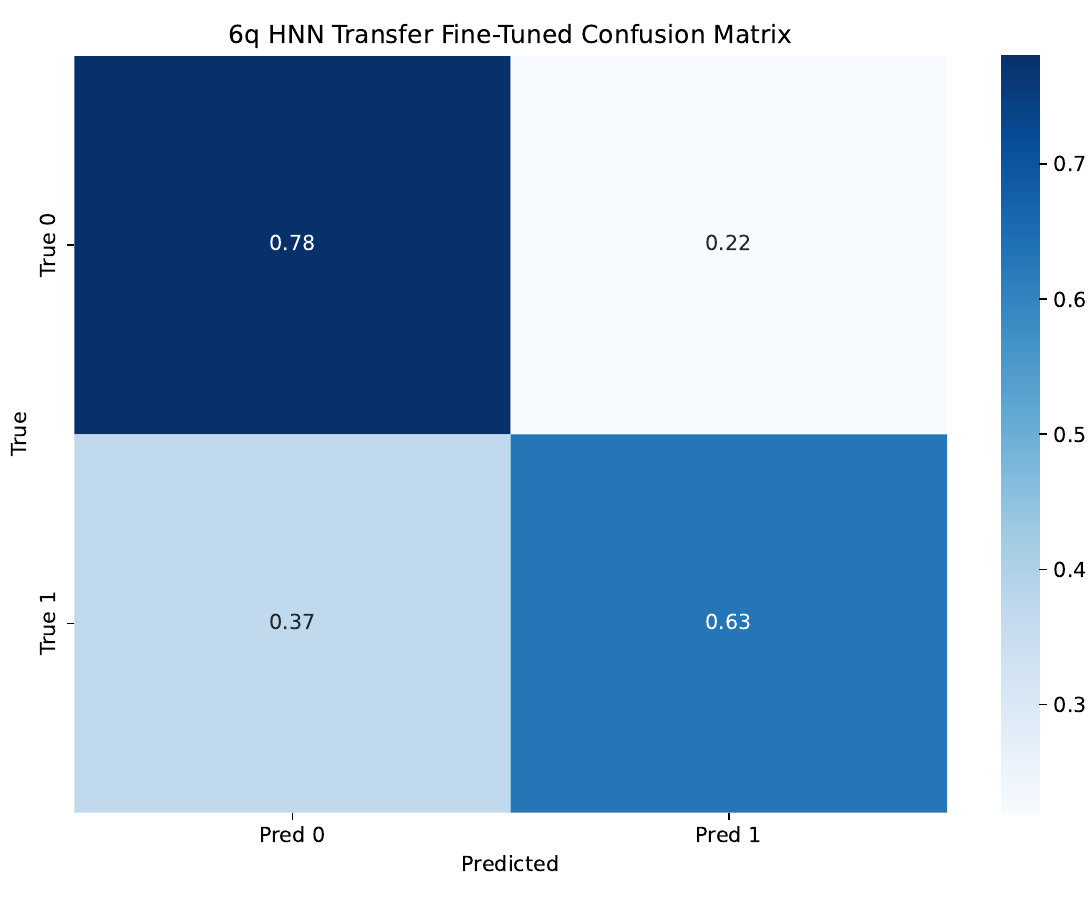}

  \caption{Average confusion matrices over 5 runs of the classical and hybrid neural network with 6-qubit quantum layer, after transfer learning on the SMS spam dataset.}
  \label{fig:trans1}
\end{figure}

\begin{figure}[H]
  \centering
  \includegraphics[scale=.55]{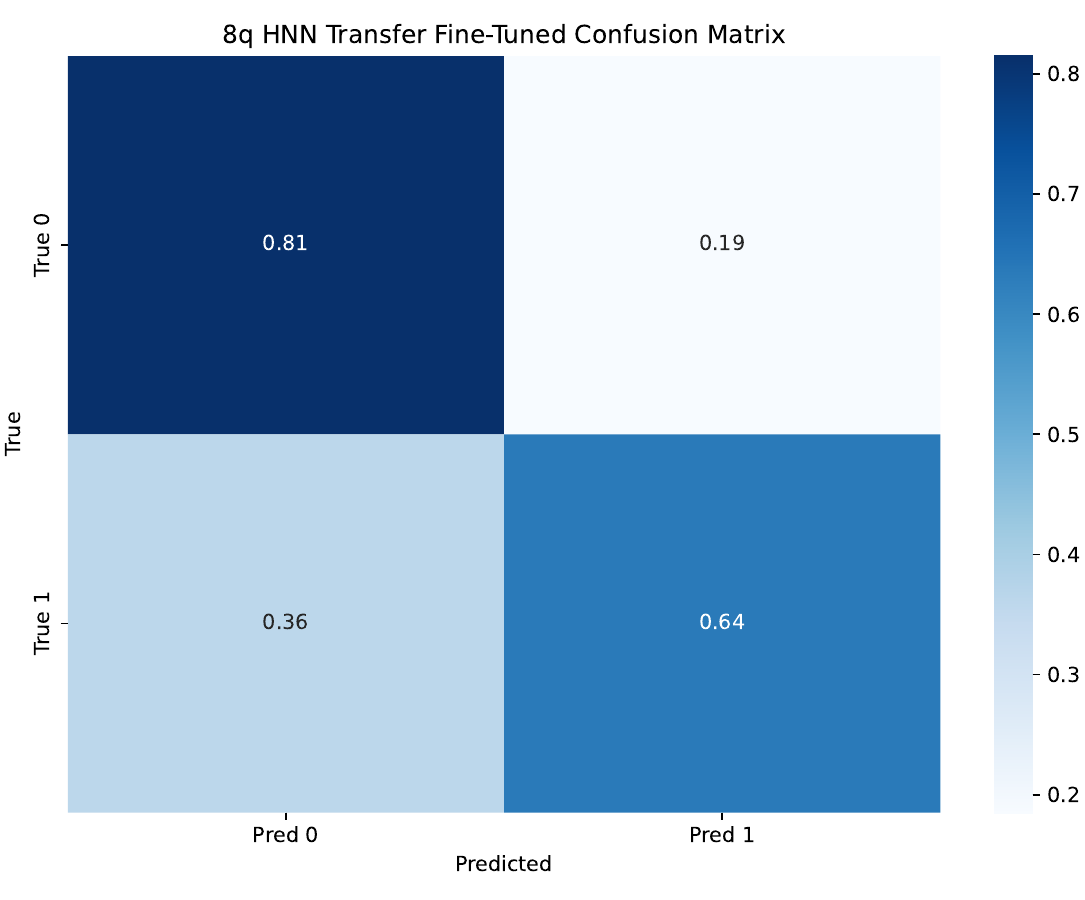}
  \includegraphics[scale=.55]{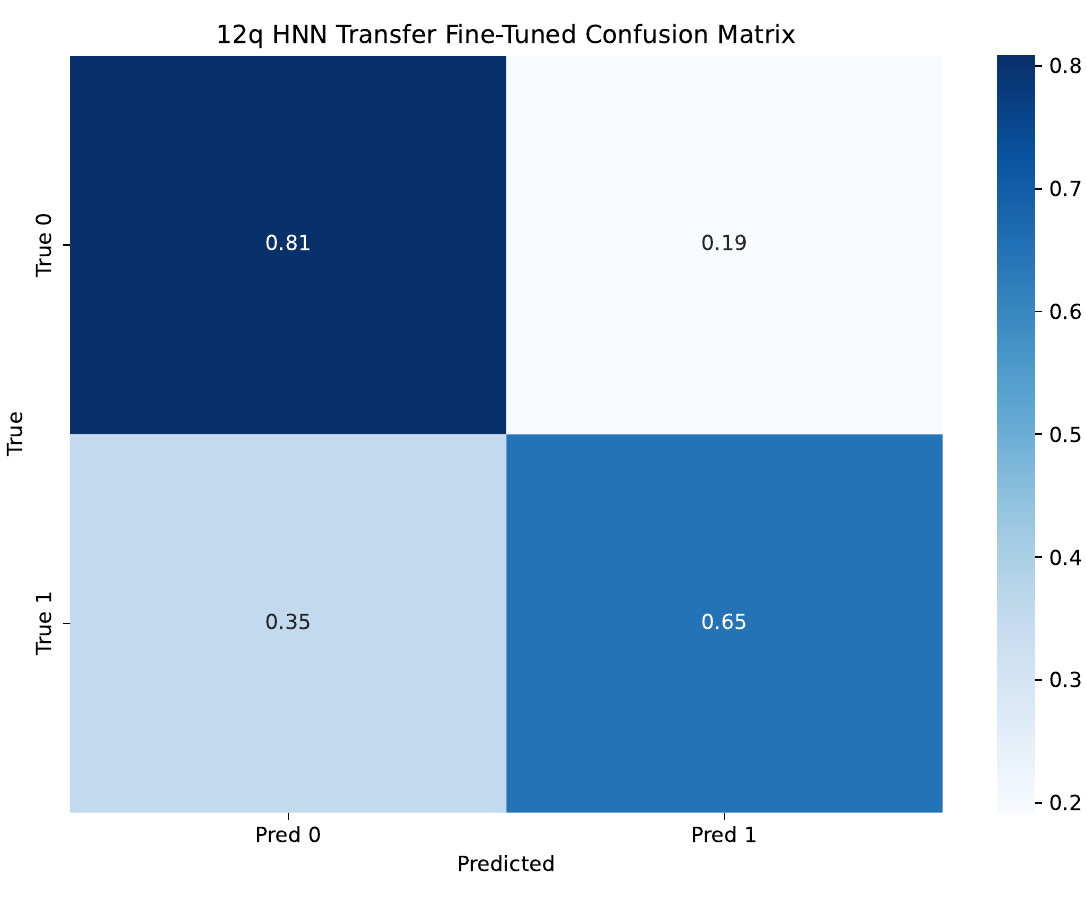}
  \caption{Average confusion matrices over 5 runs of the hybrid neural networks with 8- and 12-qubit quantum layer, after transfer learning on the SMS spam dataset.}
  \label{fig:trans2}
\end{figure}

Average confusion matrices over $5$ runs are shown in Fig. \ref{fig:trans1} and \ref{fig:trans2}, with numeric labels $0$ for ham and $1$ for spam.

The confusion matrices reveal that the performance improvement in hybrid models is primarily due to better classification of the majority class (ham), with true negative rates increasing from $66\%$ for the classical model to $81\%$ for the $12$-qubit hybrid. Performance on the minority class (spam) remains relatively stable, indicating that while false positives are reduced, recall for spam messages does not significantly improve. This asymmetry may arise from class imbalance or domain shift introduced during pretraining.

\section{Conclusions}\label{sec:con}

In this work, classical and hybrid quantum-classical neural networks were developed and evaluated for sentiment analysis tasks. The classical model, composed of fully connected layers with ReLU activations and dropout, provided a stable and reliable baseline, achieving high mean accuracy with low variance. Hybrid neural networks were constructed by embedding PQCs within classical architectures, with quantum layers of 6, 8, and 12 qubits. Classical features were encoded via angle embedding, qubits were entangled through controlled gates, and expectation values of Pauli-$Z$ operators were measured to feed subsequent classical layers. 
All quantum components were executed in a noiseless classical simulation environment.

Experiments on the COVID-related sentiment dataset showed that smaller HNNs (6- and 8-qubit) exhibited higher variability and slightly lower mean accuracies, while the 12-qubit model approached the performance of the classical baseline, highlighting the trade-off between circuit expressivity and trainability. Training dynamics indicated that the classical network converged rapidly but overfit early, whereas hybrid models displayed more gradual learning, with validation loss decreasing across the full 30 epochs. Transfer learning experiments on the SMS Spam Collection Dataset confirmed that HNNs are capable of leveraging pretrained features, particularly enhancing classification of the majority class, while performance on the minority class remained stable.

These results suggest that hybrid architectures can integrate quantum feature spaces into classical deep learning pipelines, offering potential advantages in capturing complex patterns that classical models may saturate on. However, challenges such as optimization instability, increased variance for smaller circuits, and sensitivity to circuit design remain significant, especially in the NISQ setting where noise and limited qubit connectivity must be considered \cite{du2018expressive, johnson2020kernel}.

Future directions include exploring more expressive quantum ansätze, optimizing entangling patterns and circuit depth, and systematically tuning quantum-specific hyperparameters \cite{abbas2021power}. Further studies could investigate the interplay between noise and hybrid learning, drawing parallels with classical regularization techniques \cite{du2018expressive}. To fully realize the richer representational capacity discussed, future work could also extend to quantum sequential models like Quantum Long Short-Term Memory (QLSTM) \cite{chen2022quantum} or Quantum Gated Recurrent Units (QGRU) \cite{ceschini2024variational}. This would allow hybrid models to capture not only semantic features but also the temporal dynamics and long-range contextual relationships within text sequences, offering enhancement to sentiment analysis and NLP tasks.

Beyond sentiment analysis, hybrid models hold promise for high-dimensional, complex-data tasks such as financial forecasting \cite{sakuma2020application} or cybersecurity \cite{eze2025quantum}, where quantum layers may help uncover nonlinear correlations or rare but critical patterns. These perspectives highlight the potential of hybrid quantum-classical approaches to extend the capabilities of classical machine learning as quantum hardware continues to improve.

\section*{Data availability}

The COVID-19 datasets are publicly available on Kaggle at: \\
\href{https://www.kaggle.com/datasets/datatattle/covid-19-nlp-text-classification}{https://www.kaggle.com/datasets/datatattle/covid-19-nlp-text-classification}. The SMS Spam dataset is available at \href{https://www.kaggle.com/datasets/uciml/sms-spam-collection-dataset}{https://www.kaggle.com/datasets/uciml/sms-spam-collection-dataset}.

\bibliography{sn-bibliography}

\end{document}